\pdfoutput=1

\documentclass[11pt]{article}

\usepackage[preprint]{acl}

\usepackage{times}
\usepackage{latexsym}

\usepackage[T1]{fontenc}

\usepackage[utf8]{inputenc}

\usepackage{microtype}

\usepackage{inconsolata}

\usepackage{graphicx}

\usepackage{booktabs}
\usepackage{tikz}
\usepackage{pgfplots}
\usepackage{multirow}
\usetikzlibrary{positioning,calc}%
\usetikzlibrary{backgrounds}
\usetikzlibrary{decorations.fractals}
\usetikzlibrary{patterns}
 \usepackage{amsmath}
 \usepackage{amssymb}
\usepackage[table]{xcolor}
\usepackage{makecell}
\usepackage{amsthm}
\usepackage{mathtools}

\definecolor{myYellow}{rgb}{0.9,0.9,1}
\definecolor{WindowsBlue}{RGB}{2,152,219}
\definecolor{battleshipgrey}{rgb}{0.3, 0.3, 0.3}
\definecolor{brilliantrose}{rgb}{1.0, 0.33, 0.64}
\definecolor{americanrose}{rgb}{1.0, 0.01, 0.24}
\definecolor{jweigreen}{rgb}{0,0.45,0.24}
\definecolor{bluegray}{rgb}{0.1, 0.1, 0.4}
\definecolor{ao(english)}{rgb}{0.0, 0.5, 0.0}
\definecolor{blanchedalmond}{rgb}{1.0, 0.92, 0.8}
\definecolor{atomictangerine}{rgb}{1.0, 0.6, 0.4}
\definecolor{chocolate(web)}{rgb}{0.82, 0.41, 0.12}
\definecolor{bananayellow}{rgb}{1.0, 0.88, 0.21}
\definecolor{goldenbrown}{rgb}{0.6, 0.4, 0.08}
\definecolor{aliceblue}{rgb}{0.94, 0.97, 1.0}
\definecolor{beige}{rgb}{0.96, 0.96, 0.86}
\definecolor{babyblue}{rgb}{0.54, 0.81, 0.94}
\definecolor{camel}{rgb}{0.76, 0.6, 0.42}
\definecolor{cinnamon}{rgb}{0.82, 0.41, 0.12}

\usepackage[normalem]{ulem}

\title{SPRI: SVD-Partitioned Residual Initialization for Data-Constrained MoE Upcycling}

\author{
\normalfont
Weiqiao Shan$^{1}$, Ruixiang Mao$^{1}$, Yuang Li$^{2}$, Yuhao Zhang$^{3}$, Yingfeng Luo$^{1}$,\\
Tong Zheng$^{4}$, Chen Xu$^{5}$, Yucheng Qiao$^{1}$, Chunxiang Jin$^{6}$, Yi Yuan$^{6}$,\\
Jingdong Chen$^{6}$, Tong Xiao$^{1,7}$\thanks{\ \ Corresponding author.}, Jingbo Zhu$^{1,7}$ \\
$^{1}$Northeastern University, China; $^{2}$Huawei TSC, China; $^{3}$CUHK-Shenzhen, China \\
$^{4}$University of Maryland, USA; $^{5}$Harbin Engineering University, China \\ $^{6}$Inclusion AI, Ant Group; $^{7}$NiuTrans Research, China
}

\begin{document}

\maketitle
\begin{abstract}

Mixture-of-Experts (MoE) models enable efficient scaling, but training them from scratch remains prohibitively expensive. MoE upcycling mitigates this cost by converting pretrained dense models into sparse MoE models. However, existing upcycling methods typically rely on large-scale continued training and often perform poorly under data-constrained supervised adaptation, due to either homogeneous experts or overly disruptive perturbations to pretrained parameters. In this setting, effective upcycling must leverage pretrained weight structure while introducing sufficient diversity among routed experts. To this end, we propose SVD-Partitioned Residual Initialization (SPRI), which distributes SVD-partitioned residuals derived from pretrained feed-forward network (FFN) weights across routed experts, introducing controlled expert diversity grounded in pretrained spectral structure. We further introduce a two-stage training strategy to improve adaptation stability. We evaluate SPRI on multilingual speech-to-text translation, where limited supervised data challenges MoE upcycling and multiple target languages provide natural routing heterogeneity. On CoVoST2 across 15 En-to-XX directions, SPRI improves average BLEU and COMET over fully fine-tuned dense models by 2.58 and 3.32 points, respectively, and outperforms the prior best MoE upcycling baseline by 3.39 BLEU and 4.34 COMET points.

\end{abstract}
\section{Introduction}

Large language models (LLMs) with Mixture-of-Experts (MoE) architectures have emerged as a practical paradigm for scaling model capacity without proportionally increasing per-token computation, by activating only a sparse subset of parameters for each input~\cite{jiang2024mixtral,dai2024deepseekmoe,muennighoff2025olmoe}. 
However, training MoE models from scratch remains prohibitively expensive. 
MoE upcycling offers a more efficient alternative by converting a pretrained dense model into a sparse MoE model, thereby reusing dense pretrained knowledge as a strong initialization for expert-based scaling~\cite{komatsuzaki2022sparse,he2024upcycling}.

\begin{figure}[t]
  \centering
  \centerline{\includegraphics[width=0.50\textwidth]{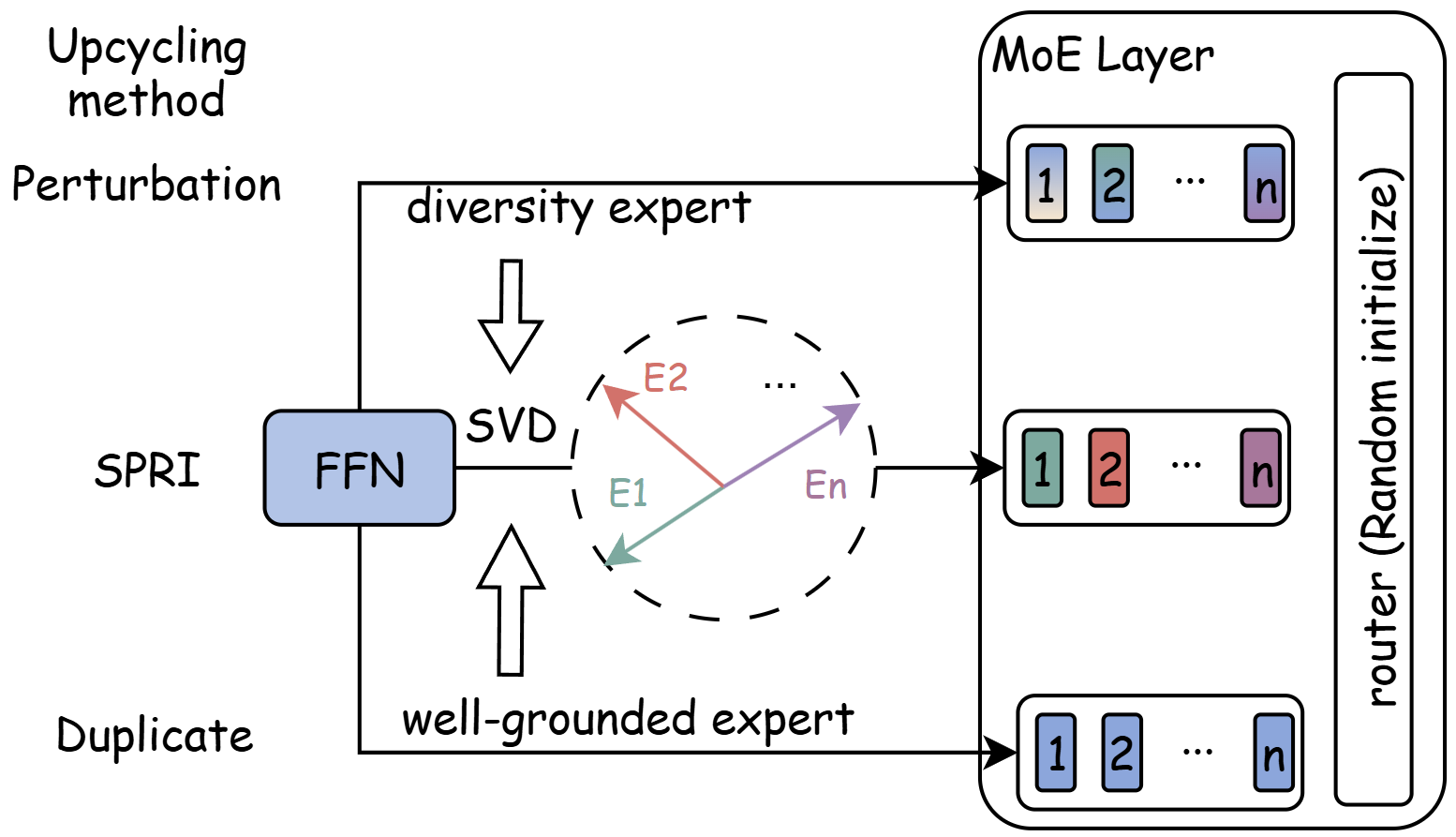}}
\caption{Overview of SVD-Partitioned Residual Initialization (SPRI). Compared with conventional upcycling methods that duplicate FFN weights or add unstructured perturbations, SPRI assigns structured spectral residuals to routed experts, yielding controlled expert diversity grounded in pretrained weight structure.}
\label{fig:simple_method}
\end{figure}

Most existing MoE upcycling methods initialize each MoE layer by duplicating the feed-forward network (FFN) of the corresponding dense Transformer block across multiple routed experts~\cite{komatsuzaki2022sparse,he2024upcycling}. 
Recent open-source MoE models further introduce random perturbations or partial re-initialization to break expert symmetry and encourage specialization~\cite{yang2024qwen2technicalreport,muennighoff2025olmoe,nakamura2025drop}. 
However, these strategies are typically paired with large-scale continued pretraining, often involving hundreds of billions or even trillions of tokens. 
This leaves an important question underexplored: \emph{can MoE upcycling still provide effective capacity expansion when only limited adaptation data is available?}

In this limited-data regime, effective MoE upcycling requires more than parameter reuse: it must leverage the internal structure of pretrained FFN weights to create diverse yet well-grounded routed experts. 
Directly copying the dense FFN provides a conservative initialization from the pretrained model, but produces nearly identical experts and limits the benefit of sparse expert modeling~\cite{nakamura2025drop}. 
As we show empirically, perturbation-based strategies can introduce expert diversity, but their unstructured modifications often make optimization less reliable in the absence of large-scale continued pretraining.
Thus, the key challenge is to construct expert heterogeneity from the pretrained model itself, rather than relying on unstructured perturbations.

To address this challenge, we propose \textbf{SVD-Partitioned Residual Initialization} (SPRI), an MoE upcycling method based on singular value decomposition (SVD). 
SPRI decomposes pretrained FFN weights and assigns structured spectral residuals from distinct subspaces to different routed experts. 
This produces bounded and interpretable expert variation while keeping the dense FFN as a functional anchor. 
We further introduce a two-stage training strategy that first adapts only the residual components and then finetunes all routed expert parameters, making MoE upcycling more stable under limited data.

We evaluate SPRI on multilingual speech-to-text translation (S2TT), a practical setting where additional expert capacity is desirable but large-scale continued pretraining is often infeasible. 
Multilingual translation benefits from expert specialization across languages and language pairs~\cite{gu2018universal,kudugunta2021exploring,zhao2024sparse}, while supervised speech-translation data~\cite{wang2020covost2} remains substantially smaller than typical continued-pretraining corpora. 
On 15 English-to-X directions in CoVoST2, SPRI improves over fully fine-tuned dense models by 2.58 BLEU and 3.32 COMET points on average, and further outperforms the prior best MoE upcycling baseline by 3.39 BLEU and 4.34 COMET points.

\section{Method}

\subsection{Mixture of Experts}

Sparse MoE layers route each token to a small subset of experts. Formally, given an input $\mathbf{x}\in\mathbb{R}^{d}$, each expert $\mathbf{E}_i$ in an MoE layer is implemented as a gated MLP:
\begin{align}
\mathbf{E}_i(\mathbf{x}) 
&= 
\mathbf{W}_i^{\mathrm{Down}}
\bigl(
\sigma(\mathbf{W}_i^{\mathrm{Gate}}\mathbf{x}) 
\odot 
\mathbf{W}_i^{\mathrm{Up}}\mathbf{x}
\bigr),
\label{eq:ffn_expert}
\end{align}
where $\odot$ denotes element-wise multiplication and $\sigma(\cdot)$ is the activation function.

At layer $l$, the router assigns token to the routed experts and activates the Top-$k$ experts in $E$ routed experts, where $k \ll E$:
\begin{align}
    \mathbf{g}^l(\mathbf{x})
    &=
    \operatorname{Softmax}
    \left(
    \left\{
    (\mathbf{W}^l_{\mathrm{router}}\mathbf{x})_i
    \right\}_{i\in \mathcal{S}_k^l(\mathbf{x})}
    \right), \notag \\
    \mathcal{S}_k^l(\mathbf{x})
    &=
    \operatorname{TopK}
    \left(
    \mathbf{W}^l_{\mathrm{router}}\mathbf{x}, k
    \right).
\end{align}

Following the recent MoE architectures, which introduce shared experts that are always activated~\citep{dai2024deepseekmoe,deepseekai2024deepseekv2}, the MoE layer is then computed as:
\begin{align}
\label{eq:shared_moe}
    \mathrm{MoE}^l(\mathbf{x})
    &=
    \sum_{i\in \mathcal{S}_k^l(\mathbf{x})} 
    g^l_i(\mathbf{x}) \mathbf{E}^l_i(\mathbf{x}) 
    + 
    \mathbf{E}^l_{\mathrm{shared}}(\mathbf{x}).
\end{align}

\subsection{Preliminary Analysis of MoE Upcycling Strategies}

We first review conventional MoE upcycling strategies and formulate them under a unified notation.

\paragraph{Naive Upcycling.}
Naive upcycling~\cite{komatsuzaki2022sparse,he2024upcycling} converts a pretrained dense model into a sparse MoE model by initializing all routed experts with the original dense MLP module and randomly initializing the router. For each expert $i\in\{1,\ldots,E\}$ in naive upcycling, the initialization is:
\begin{align}
\label{eq:naive_upcycling}
\mathbf{W}^{(\cdot)}_{i} = \mathbf{W}^{(\cdot)}_{\mathrm{MLP}},
\end{align}
where $(\cdot)\in\{\mathrm{Gate},\mathrm{Up},\mathrm{Down}\}$.

\paragraph{Noise Upcycling.}
Noise upcycling~\cite{yang2024qwen2technicalreport}\footnote{Since the Qwen2 technical report does not fully specify the perturbation procedure, we follow the descriptions in~\cite{muennighoff2025olmoe,nakamura2025drop}.} breaks the symmetry among duplicated experts by adding random perturbations to a subset of copied weights. For each expert $i$, the initialized weight matrix is:
\begin{align}
\label{eq:noise_upcycling}
\mathbf{W}^{(\cdot)}_{i}
=
\mathbf{W}^{(\cdot)}_{\mathrm{MLP}}
+
(\mathbf{1}-\mathbf{M}^{(\cdot)}_{i}) 
\odot 
\boldsymbol{\epsilon}^{(\cdot)}_{i},
\end{align}
where $\mathbf{M}^{(\cdot)}_{i}$ is a binary mask indicating the copied parameters, and each entry of $\boldsymbol{\epsilon}^{(\cdot)}_{i}$ is independently sampled from $\mathcal{N}(0,\sigma^2)$.

\begin{figure*}[t]
  \centering
  \centerline{\includegraphics[width=0.88\textwidth]{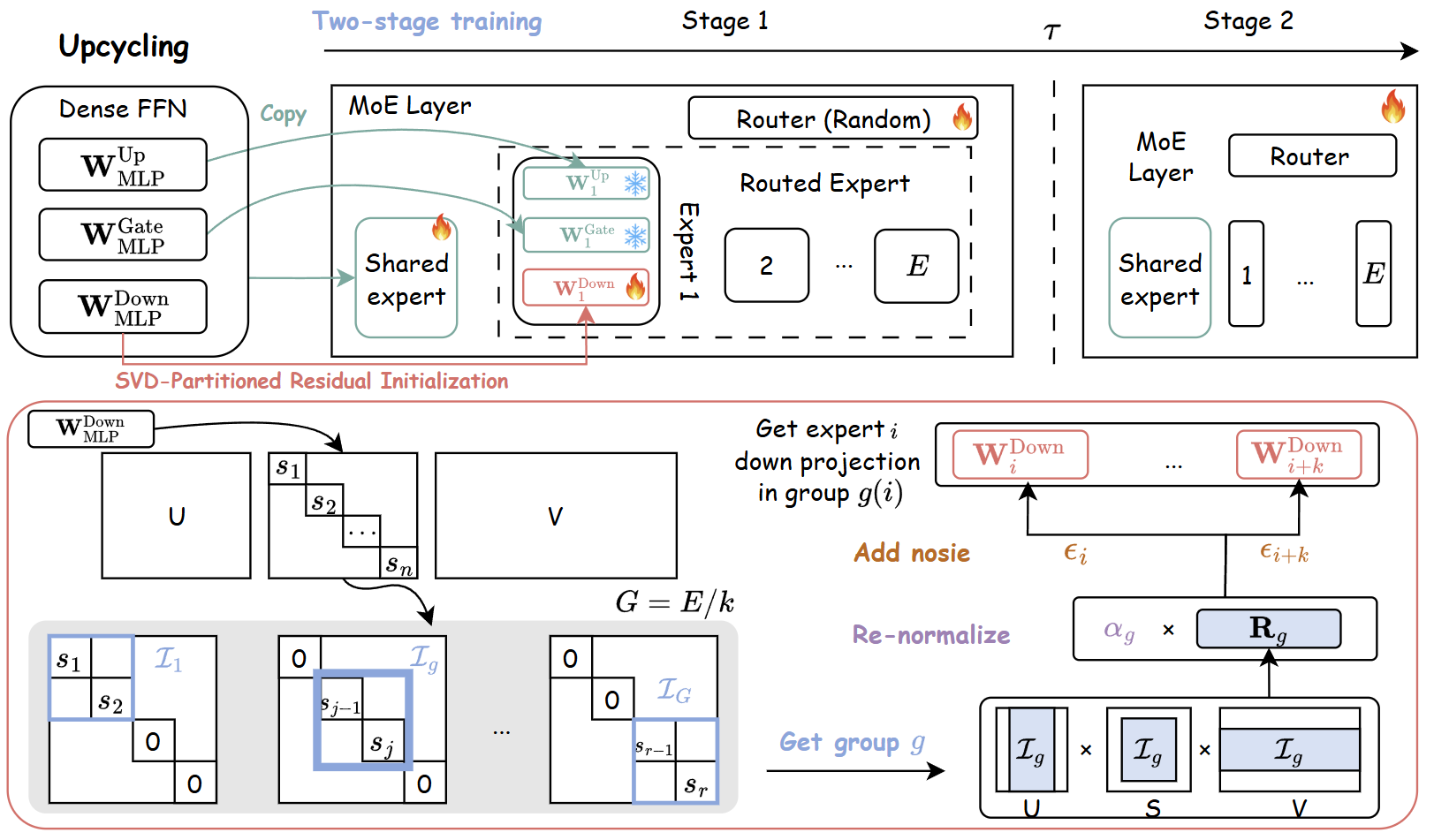}}
\caption{Pipeline of SVD-Partitioned Residual Upcycling. The shared expert copies the dense MLP to preserve pretrained behavior, while routed experts receive norm-controlled residual down projections constructed from disjoint SVD partitions.}
\label{fig:final-result}
\end{figure*}

\paragraph{Drop Upcycling.}
Drop upcycling~\cite{nakamura2025drop} promotes expert diversity by partially reinitializing the MLP parameters of duplicated experts. It randomly selects a subset of indices along the intermediate dimension for each copied expert and reinitializes the corresponding rows or columns of the MLP weight matrices, while leaving the remaining parameters unchanged:
\begin{align}
\label{eq:drop_upcycling}
\mathbf{W}^{(\cdot)}_{i}
=
\mathbf{M}^{(\cdot)}_{i} \odot \mathbf{W}^{(\cdot)}_{\mathrm{MLP}}
+
\bar{\mathbf{M}}^{(\cdot)}_{i} \odot \mathbf{R}^{(\cdot)}_{i},
\end{align}
where $\mathbf{M}^{(\cdot)}_{i}$ denotes a structured binary mask induced by the selected intermediate dimensions, 
$\bar{\mathbf{M}}^{(\cdot)}_{i}=\mathbf{1}-\mathbf{M}^{(\cdot)}_{i}$ is its complementary mask, 
and $\mathbf{R}^{(\cdot)}_{i}$ denotes newly initialized parameters.

\paragraph{Branch-Train-MiX (BTX).}
BTX~\cite{sukhbaatar2024branch} first trains multiple dense models on different domain-specific datasets and then merges their MLP module as routed experts, while averaging the remaining shared parameters:
\begin{align}
\label{eq:btx_upcycling}
\mathbf{W}^{(\cdot)}_{i}
=
\mathbf{W}^{(\cdot)}_{\mathrm{MLP}, i},
\end{align}
where $\mathbf{W}^{(\cdot)}_{\mathrm{MLP}, i}$ denotes the corresponding MLP weight from the $i$-th branch-trained dense model.

\paragraph{Preservation and Diversity Analysis.}

\begin{table}[t]
    \centering
    \small
    \resizebox{\linewidth}{!}{\begin{tabular}{l|c c c}
        \toprule
        Method & KL $\downarrow$ & Diversity $\uparrow$ & BLEU $\uparrow$ \\
        \midrule
        Dense FT & 0 & - & 19.66 \\
        Naive Upcycling & 1.64 & 0.01\% & 17.19 \\
        Noise Upcycling & 1.64 & 2\% & 18.36 \\
        Drop Upcycling & 4.93 & 99.99\% & 18.85 \\
        SPRI (ours) & 0.12 & 88.02\% & 22.24 \\
        \bottomrule
    \end{tabular}}
    \caption{Preservation and diversity comparison across MoE upcycling methods. Dense FT denotes the fully fine-tuned dense model. KL measures the output-distribution deviation between each upcycled model and the original dense model, diversity measures the discrepancy among routed experts, and BLEU denotes the average BLEU score across 15 En$\rightarrow$XX translation directions. Statistical details in Appendix~\ref{app:diagnostic_details}.}
    \label{table:diagnostic}
\end{table}

We analyze how initialization impacts knowledge preservation and expert diversity in MoE upcycling.\footnote{BTX is excluded from this analysis because it requires additional branch training and is not a direct single-model upcycling method.}

As shown in Table~\ref{table:diagnostic}, MoE architectures with higher diversity generally achieve higher BLEU scores, suggesting that expert diversity is beneficial for MoE architectures in our experimental setting. However, diversity alone is insufficient, all prior MoE upcycling baselines underperform the fully fine-tuned dense model with larger output KL divergence. This indicates that, when large-scale pretraining data is unavailable, effective MoE upcycling should introduce expert diversity while preserving the original dense model behavior as much as possible.

\subsection{SVD-Partitioned Residual Upcycling}

Motivated by the above analysis, we propose SPRI, which follows a dense-anchored residual design. The shared expert preserves the pretrained dense MLP module, while routed experts introduce small, structured residual corrections. This design separates knowledge preservation from expert diversification, allowing the initialized MoE to remain close to the dense model while providing diversity among routed experts.

Instead of copying or perturbing the full MLP parameters for each routed expert, SPRI copies the gate and up projections to preserve the dense nonlinear feature transformation, and introduces expert-specific differences only through small SVD-partitioned residuals in the down projection. By controlling the residual magnitude, routed experts start as weak but structured corrections.

\paragraph{Dense-anchored shared expert.}

In specific MoE layer, the shared expert provides the anchor by fully copying the pretrained dense MLP module:
\begin{align}
\mathbf{W}^{(\cdot)}_{\mathrm{shared}} 
= 
\mathbf{W}^{(\cdot)}_{\mathrm{MLP}},
\end{align}
where $(\cdot)\in\{\mathrm{Gate},\mathrm{Up},\mathrm{Down}\}$.

\paragraph{SVD-partitioned routed residuals.}

In each routed expert, the gate and up projections are initialized as:
\begin{align}
\mathbf{W}^{(\mathrm{Gate})}_{i} &= \mathbf{W}^{(\mathrm{Gate})}_{\mathrm{MLP}}, \notag \\
\mathbf{W}^{(\mathrm{Up})}_{i} &= \mathbf{W}^{(\mathrm{Up})}_{\mathrm{MLP}},
\quad i=1,\ldots,E.
\end{align}

The down projection is then used to introduce structured diversity among routed experts. Specifically, let the dense down-projection matrix be decomposed by singular value decomposition:
\begin{align}
\mathbf{W}^{(\mathrm{Down})}_{\mathrm{MLP}}
=
\mathbf{U}\operatorname{diag}(\mathbf{s})\mathbf{V}^{\top},
\end{align}
where $\mathbf{s}=(s_1,\ldots,s_r)$ contains the singular values in descending order.
We use the routing Top-$k$ as the expert group size and set the number of residual groups to $G=E/k$.
For expert $i$, we denote its group assignment by $g(i)=\lceil i/k\rceil$, where $i\in\{1,\ldots,E\}$ and $g(i)\in\{1,\ldots,G\}$.

Each group is associated with one spectral residual matrix. Specifically, we split the ordered singular-component indices $\{1,\ldots,r\}$ into $G$ contiguous blocks $\{\mathcal{I}_g\}_{g=1}^{G}$. For group $g \in \{1,\ldots,G\}$, the corresponding residual matrix is reconstructed as:
\begin{align}
\label{eq:svd_partition}
\mathbf{R}^{(\mathrm{Down})}_{g}
=
\mathbf{U}_{[:, \mathcal{I}_g]}
\operatorname{diag}\bigl((s_j)_{j\in\mathcal{I}_g}\bigr)
\mathbf{V}_{[:, \mathcal{I}_g]}^{\top}.
\end{align}
The down projection of routed expert $i \in \{1,\ldots,E\}$ is initialized from the residual matrix of its group:
\begin{align}
\label{eq:svd_partitioned_residual}
\mathbf{W}^{(\mathrm{Down})}_{i}
=
\alpha_{g(i)} \mathbf{R}^{(\mathrm{Down})}_{g(i)}
+
\boldsymbol{\epsilon}_{i},
\end{align}
where $\boldsymbol{\epsilon}_{i}$ is a small expert-specific perturbation that breaks symmetry among experts in the same group.

While groups contain larger singular values tend to yield residual matrices $\mathbf{R}^{(\mathrm{Down})}_{g(i)}$ with larger Frobenius norms. We introduce a group-wise scaling coefficient to re-normalize each residual matrices:
\begin{align}
\alpha_g
=
\frac{
\rho \left\|\mathbf{W}^{(\mathrm{Down})}_{\mathrm{MLP}}\right\|_{F}
}{
\left\|\mathbf{R}^{(\mathrm{Down})}_{g}\right\|_{F} + \delta
},
\end{align}
where $\rho$ controls the relative magnitude of each routed residual and $\delta$ is a small constant for numerical stability. 
A small $\rho$ makes the initial MoE rely primarily on the shared expert as the preservation anchor.

\paragraph{Residual MoE interpretation.}

As a result, the initialized MoE layer can be viewed as the pretrained dense FFN plus a small routed residual term:
\begin{align}
\mathrm{SPRI}^{l}(\mathbf{x})
&=
\mathbf{E}^{l}_{\mathrm{shared}}(\mathbf{x})
+
\sum_{i\in \mathcal{S}_k^l(\mathbf{x})}
g_i^l(\mathbf{x})\Delta \mathbf{E}^{l}_i(\mathbf{x}) \notag \\
\Delta \mathbf{E}^{l}_i(\mathbf{x})
&=
\mathbf{W}^{(\mathrm{Down})}_{i}
\left(
\sigma(\mathbf{W}^{(\mathrm{Gate})}_{\mathrm{FFN}}\mathbf{x})
\odot
\mathbf{W}^{(\mathrm{Up})}_{\mathrm{FFN}}\mathbf{x}
\right)
\end{align}

\subsection{Training Strategy}\label{sec:training_strategy}

\paragraph{Two-stage training}
After initialization, we apply a two-stage update schedule to routed experts, to exploit the expert diversity.

During the first $\tau T$ training steps, only $\{\mathbf{W}^{(\mathrm{Down})}_{i}\}_{i=1}^{E}$ and the router are updated, while the copied gate and up projections are frozen. 
Afterward, $\{\mathbf{W}^{(\mathrm{Gate})}_{i}, \mathbf{W}^{(\mathrm{Up})}_{i}, \mathbf{W}^{(\mathrm{Down})}_{i}\}_{i=1}^{E}$ are optimized jointly with the router. 
Here, $T$ is the total number of training steps and $\tau$ is the freezing ratio.

\paragraph{Training objective}
We train all MoE models with the standard MoE training objective, consisting of the task cross-entropy loss and the auxiliary routing losses commonly used in sparse MoE models~\cite{fedus2022switch,zoph2022stmoe}. The final objective is
\begin{align}
\label{eq:training_objective}
\mathcal{L}
=
\mathcal{L}_{\mathrm{CE}}
+
\lambda_{\mathrm{lb}}\mathcal{L}_{\mathrm{lb}}
+
\lambda_{z}\mathcal{L}_{z},
\end{align}

where $\mathcal{L}_{\mathrm{CE}}$ is the task cross-entropy loss, $\mathcal{L}_{\mathrm{lb}}$ is balance loss which encourages balanced expert utilization, and $\mathcal{L}_{z}$ is z-loss which regularizes router logits for stable training. The coefficients $\lambda_{\mathrm{lb}}$ and $\lambda_z$ control the strengths of the corresponding auxiliary losses.

\section{Experimental Setups}

\subsection{Datasets and Evaluation Metrics}

We train our method on the S2TT task using the CoVoST 2~\cite{wang2020covost2} and Europarl-ST~\cite{iranzo-sanchez2020europarl}, both large-scale multilingual speech translation datasets. 
We focus on the English-to-X setting to prevent the capability of the audio encoder from interfering with our experiments. We evaluate on all 15 target-language directions in the CoVoST 2 dataset. 
After preprocessing, the training split contains about 6.9k hours of speech.

We evaluate our method on CoVoST 2. We report translation quality using sacreBLEU~\cite{post2018call} and COMET~\cite{rei2020comet}. We follow the evaluation setting of Qwen2-Audio~\cite{chu2024qwen2audio} to compute all BLEU scores. For COMET, we use the \texttt{Unbabel/wmt22-comet-da}\footnote{https://huggingface.co/Unbabel/wmt22-comet-da}. 
Additional details on preprocessing, data loading, and evaluation settings are provided in Appendix~\ref{app:experimental_details}.

\subsection{Model Architecture}

We conduct our main experiments using Qwen3-ASR-0.6B~\cite{Qwen3-ASR}\footnote{The ``0.6B'' refers to the language model decoder (596M parameters with tied embeddings). Including the audio encoder, the full model has approximately 0.78B parameters, rounded to 0.7B in Table~\ref{tab:en2xx}.}, which consists of an acoustic encoder, a lightweight adapter that aligns acoustic features with the textual embedding space, and a decoder-only language model. Its moderate scale makes large-scale continued pretraining infeasible, which matches the limited-data scenario considered in this work.
To construct sparse MoE variants, we convert the dense FFN modules in the language decoder into MoE layers at a regular interval of every four decoder layers.
Each MoE layer contains 8 routed experts and 1 shared expert, and activates the Top-2 routed experts per token during both training and inference.
For all MoE upcycling methods, we use the same shared-expert configuration, where the shared expert is initialized by copying the corresponding dense FFN. Different upcycling methods mainly differ in how they initialize the routed experts.

\begin{table*}[t]
\centering
\resizebox{\textwidth}{!}{\begin{tabular}{l|c|c c c c c c c c c c c c c c c c}
\hline
\rowcolor{black!10} & & \multicolumn{2}{c}{\textbf{Tr}} & \multicolumn{2}{c}{\textbf{Ta}} & \multicolumn{2}{c}{\textbf{Ja}} & \multicolumn{2}{c}{\textbf{Sl}} & \multicolumn{2}{c}{\textbf{Lv}} & \multicolumn{2}{c}{\textbf{Fa}} & \multicolumn{2}{c}{\textbf{Et}} & \multicolumn{2}{c}{\textbf{De}}\\ 
\rowcolor{black!10} \multirow{-2}{*}{Model} & \multirow{-2}{*}{Param} & COMET & BLEU & COMET & BLEU & COMET & BLEU & COMET & BLEU & COMET & BLEU & COMET & BLEU & COMET & BLEU & COMET & BLEU\\ 
\hline
Dense FT & 0.7B & \textbf{77.28} & \textbf{12.46} & 75.43 & 8.79 & \textbf{86.01} & \textbf{30.36} & 68.82 & 14.69 & 66.12 & 10.27 & 68.94 & 11.17 & 65.38 & 10.67 & \textbf{80.75} & \textbf{28.52} \\
LoRA & 0.4B & 74.89 & 11.02 & \textbf{79.20} & \textbf{10.95} & 84.73 & 28.93 & \textbf{68.86} & \textbf{14.83} & \textbf{69.31} & \textbf{11.77} & \textbf{71.05} & \textbf{12.53} & \textbf{67.09} & \textbf{10.68} & 78.25 & 24.92 \\
Naive Upcycling & 1.3B & 73.94 & 10.36 & 75.18 & 8.33 & 84.41 & 28.35 & 65.51 & 12.43 & 64.18 & 8.85 & 68.01 & 10.55 & 62.47 & 8.56 & 77.99 & 24.60 \\
Drop Upcycling & 1.3B & 75.94 & 11.56 & 74.60 & 8.46 & 85.34 & 29.14 & 67.63 & 14.26 & 65.28 & 9.81 & 67.35 & 10.41 & 64.29 & 10.17 & 79.45 & 27.23 \\
Noise Upcycling & 1.3B & 75.37 & 10.98 & 73.87 & 8.11 & 85.12 & 28.77 & 66.91 & 13.63 & 64.51 & 9.53 & 66.76 & 10.18 & 63.38 & 9.92 & 78.89 & 26.58 \\
BTX & 1.3B & 74.58 & 10.88 & 72.93 & 7.75 & 84.80 & 28.35 & 66.09 & 13.60 & 64.33 & 9.64 & 66.56 & 10.09 & 63.08 & 10.28 & 78.47 & 26.39 \\
\hline
SPRI & 1.3B & 78.51 & 13.58 & 78.93 & 11.02 & 86.24 & 30.69 & 72.58 & 17.66 & 71.05 & 12.98 & 72.26 & 13.24 & 70.71 & 13.62 & 80.98 & 29.04 \\
SPRI 2 stage & 1.3B & \textbf{78.85} & \textbf{13.83} & \textbf{79.80} & \textbf{11.52} & \textbf{86.28} & \textbf{30.80} & \textbf{74.25} & \textbf{19.13} & \textbf{73.45} & \textbf{14.20} & \textbf{73.08} & \textbf{13.75} & \textbf{71.90} & \textbf{14.14} & \textbf{81.03} & \textbf{29.16} \\
\hline
\end{tabular}}

\resizebox{\textwidth}{!}{\begin{tabular}{l|c|c c c c c c c c c c c c c c|c c}
\hline
\rowcolor{black!10} & & \multicolumn{2}{c}{\textbf{Zh}} & \multicolumn{2}{c}{\textbf{Mn}} & \multicolumn{2}{c}{\textbf{Sv}} & \multicolumn{2}{c}{\textbf{Cy}} & \multicolumn{2}{c}{\textbf{Ca}} & \multicolumn{2}{c}{\textbf{Ar}} & \multicolumn{2}{c}{\textbf{Id}} & \multicolumn{2}{c}{\textbf{Avg}} \\
\rowcolor{black!10} \multirow{-2}{*}{Model} & \multirow{-2}{*}{Param} & COMET & BLEU & COMET & BLEU & COMET & BLEU & COMET & BLEU & COMET & BLEU & COMET & BLEU & COMET & BLEU & COMET & BLEU \\ 
\hline
Dense FT & 0.7B & \textbf{84.40} & \textbf{42.42} & 67.62 & 4.37 & \textbf{76.58} & \textbf{26.71} & 67.82 & 16.40 & \textbf{75.36} & \textbf{26.26} & \textbf{80.07} & \textbf{18.41} & \textbf{85.92} & \textbf{33.42} & 75.10 & \textbf{19.66} \\
LoRA & 0.4B & 82.82 & 38.76 & \textbf{74.24} & \textbf{7.11} & 73.38 & 23.05 & \textbf{70.73} & \textbf{18.42} & 74.42 & 25.49 & 79.06 & 17.32 & 82.73 & 27.97 & \textbf{75.38} & 18.92 \\
Naive Upcycling & 1.3B & 82.44 & 37.81 & 68.01 & 4.48 & 72.58 & 22.46 & 64.93 & 13.48 & 72.50 & 22.80 & 78.89 & 17.07 & 82.53 & 27.74 & 72.91 & 17.19 \\
Drop Upcycling & 1.3B & 83.84 & 40.88 & 67.54 & 4.52 & 74.32 & 25.14 & 67.47 & 16.20 & 74.51 & 25.61 & 78.82 & 17.25 & 84.87 & 32.08 & 74.08 & 18.85 \\
Noise Upcycling & 1.3B & 83.58 & 40.40 & 66.64 & 4.22 & 73.52 & 24.59 & 66.82 & 15.71 & 73.84 & 24.73 & 78.50 & 16.64 & 84.55 & 31.37 & 73.48 & 18.36 \\
BTX & 1.3B & 83.17 & 39.69 & 66.70 & 4.31 & 73.22 & 24.46 & 66.78 & 15.63 & 73.55 & 24.70 & 77.88 & 16.53 & 84.13 & 31.38 & 73.09 & 18.24 \\
\hline
SPRI & 1.3B & 84.56 & 42.85 & 73.77 & \textbf{7.11} & 77.72 & 28.15 & 73.36 & 22.22 & 77.10 & 29.30 & 80.41 & 18.80 & 86.12 & 33.89 & 77.62 & 21.61 \\
SPRI 2 stage & 1.3B & \textbf{84.57} & \textbf{43.04} & \textbf{75.57} & \textbf{7.94} & \textbf{77.98} & \textbf{28.52} & \textbf{75.05} & \textbf{24.06} & \textbf{77.81} & \textbf{30.42} & \textbf{80.46} & \textbf{19.03} & \textbf{86.26} & \textbf{34.14} & \textbf{78.42} & \textbf{22.24} \\
\hline
\end{tabular}}
\caption{Results on CoVoST2 English-to-X speech-to-text translation. We report COMET and sacreBLEU for each target language and their average over 15 directions. Dense FT denotes the fully fine-tuned dense model.}
\label{tab:en2xx}
\end{table*}

\subsection{Training}

During training, we keep all audio inputs resampled to 16 kHz and set the maximum sequence length to 768. 
We train each model for 1 epoch with a global batch size of 1024. Optimization is performed with AdamW using a peak learning rate of $2\times10^{-5}$, a linear learning rate schedule, and a warmup ratio of 0.02 with bf16 precision.

For MoE models, we use the same router regularization losses and coefficients, with $\lambda_{\mathrm{lb}}=10^{-2}$ and $\lambda_{z}=10^{-3}$ across all upcycling methods. 

For SPRI, we apply the two-stage training strategy described in Section~\ref{sec:training_strategy}, with $\tau=0.1$, and we select $\rho=10^{-3}$, $\delta=10^{-12}$ for all MoE models. All experiments are conducted using 8 NVIDIA H800 GPUs with 80GB memory each.

\section{Results}

\subsection{Main Results}

Table~\ref{tab:en2xx} reports the main results on CoVoST2 English-to-X S2TT dataset. 
Conventional MoE upcycling baselines fail to consistently benefit from the increased parameter count in the data-constrained setting. 
Although these methods expand the model from 0.7B to 1.3B parameters, Naive, Drop, Noise, and BTX Upcycling all underperform Dense FT on average. 
This suggests that effective upcycling requires more than directly reusing or perturbing dense FFN parameters.

Among conventional upcycling methods, Drop and Noise improve over Naive Upcycling, indicating that expert diversity is more useful than homogeneous expert duplication. 
However, their performance remains below Dense FT, showing that unstructured perturbations are insufficient for reliable expert specialization. 
BTX also fails to close this gap, suggesting that decoupling expert construction from router adaptation is less effective than jointly optimizing routed experts and the router.

SPRI consistently outperforms all baselines. 
With two-stage training, it achieves 78.42 COMET and 22.24 BLEU, improving over Dense FT by +3.32 COMET and +2.58 BLEU and over the best conventional MoE upcycling baseline by +4.34 COMET and +3.39 BLEU. 
The single-stage variant also outperforms all baselines, indicating that the main gain comes from SVD-partitioned residual initialization, while two-stage training further improves performance. 
These results demonstrate that SPRI better balances pretrained knowledge preservation and expert diversification than homogeneous copying or unstructured perturbations.

SPRI improves across all 15 translation directions, with smaller gains on high-resource languages such as Zh, De, and Ja. 
We conjecture that the pretrained dense model has already learned strong representations for these languages from higher-resource pretraining data, so the CoVoST2 fine-tuning signal provides relatively limited additional benefit. 
The degradation of conventional upcycling methods on these directions is consistent with this view: when the dense initialization is already strong, disrupting pretrained knowledge can be more harmful than the capacity gain from MoE upcycling.

To further verify the effectiveness of SPRI, we conduct a series of controlled ablation studies. 
Rather than tuning each component for the best possible score, these experiments aim to isolate the contribution of individual design choices. 
Unless otherwise specified, we vary one factor at a time while keeping the remaining settings fixed.

\subsection{Ablation on SVD-Partitioned Routed Residuals}

\begin{table}[htp]
\centering
\small
\begin{tabular}{lcc}
\toprule
Variant & Avg. COMET & Avg. BLEU \\
\midrule
\multicolumn{3}{c}{\textit{Down-projection initialization}} \\
\midrule
Copy            & 75.16 & 19.64 \\
Noise           & 75.11 & 19.59 \\
Drop            & 74.23 & 18.94 \\
SVD-Partitioned & \textbf{78.42} & \textbf{22.24} \\
\midrule
\multicolumn{3}{c}{\textit{Shared expert}} \\
\midrule
w/ shared expert  & \textbf{78.42} & \textbf{22.24} \\
w/o shared expert & 78.28 & \textbf{22.24} \\
\bottomrule
\end{tabular}
\caption{
Ablation on SVD-partitioned routed residuals. 
The first block keeps the SPRI framework fixed and varies only the routed down-projection initialization. 
The second block removes the shared expert to test whether SPRI relies on the dense preservation anchor. All direction detailed and experimental settings are provided in Appendix~\ref{app:down_projection_initialization_full_results}.
}
\label{tab:down_residual_construction}
\end{table}

We further verify whether the gains of SPRI come from the SVD-partitioned routed residuals themselves. 
Table~\ref{tab:down_residual_construction} reports two ablation experiments.
In the first part, we only replace the routed down-projection initialization.

Copy, noise, and drop initialization achieve 19.64, 19.59, and 18.94 BLEU, respectively, while SVD-partitioned residuals achieve 22.24 BLEU and 78.42 COMET. 
This shows that the improvement is not simply due to applying upcycling only to the down projection or using the same training strategy. 
Instead, constructing routed experts from structured spectral directions is important for inducing useful expert diversity grounded in the pretrained FFN weights.

The second part examines whether SPRI depends on the shared expert. 
Removing the shared expert gives the same BLEU as the default setting and only slightly decreases COMET from 78.42 to 78.28. 
This indicates that the main gain comes from SVD-partitioned routed residuals rather than from the shared expert alone. 
The shared expert can serve as a preservation anchor, but it is not necessary for SPRI to remain effective. 

\subsection{Parameter-Matched Scaling via MoE Upcycling}

We further compare SPRI-based MoE upcycling with dense models at a similar total parameter scale. 
As shown in Table~\ref{tab:parameter_matched_dense_comparison}, SPRI-3B achieves 80.57 COMET and 24.43 BLEU, substantially outperforming both \texttt{Qwen2.5-Omni-3B} and its fine-tuned variant. 
Compared with \texttt{Qwen2.5-Omni-3B FT}, SPRI-3B improves the average score by +3.98 COMET and +4.48 BLEU.

These results suggest that, under data-constrained S2TT adaptation, expanding a smaller pretrained model through MoE upcycling can be more effective than directly fine-tuning a parameter-matched dense model. 
Since the upcycled MoE activates only a subset of experts per token, it also offers a more computation-efficient scaling path than dense models with a similar total parameter count.

\begin{table}[htp]
\centering
\small
\begin{tabular}{lccc}
\toprule
Model & Avg. COMET & Avg. BLEU \\
\midrule
\texttt{Qwen2.5-Omni-3B} & 68.07 & 13.86 \\
\texttt{Qwen2.5-Omni-3B FT} & 76.59 & 19.95 \\
SPRI-3B & \textbf{80.57} & \textbf{24.43} \\
\bottomrule
\end{tabular}
\caption{
Parameter-matched comparison between SPRI-based MoE upcycling and dense models. 
At a comparable total parameter scale, SPRI-3B outperforms both \texttt{Qwen2.5-Omni-3B} and its fine-tuned variant on average COMET and BLEU. Full per-language results are provided in Appendix~\ref{app:parameter_matched_full_results}.
}
\label{tab:parameter_matched_dense_comparison}
\end{table}

\subsection{Sensitivity to the Unfreezing Ratio}

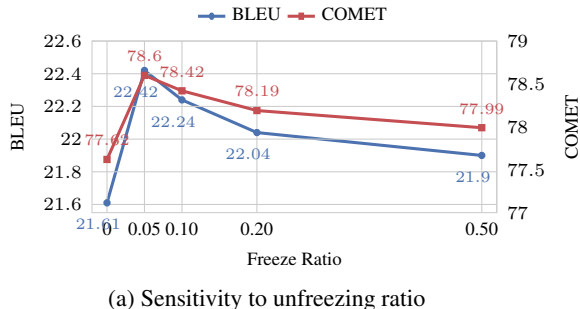
\begin{figure}[htp]
\centerline{\pgfplotsset{compat=1.17}

\definecolor{icmlblue}{RGB}{76,114,176}
\definecolor{icmlorange}{RGB}{221,132,82}
\definecolor{icmlgreen}{RGB}{85,168,104}
\definecolor{icmlred}{RGB}{196,78,82}
\definecolor{icmlpurple}{RGB}{129,114,178}
\definecolor{icmlbrown}{RGB}{147,120,96}

\begin{tikzpicture}
\def\barbias{0.5em}

\begin{scope}[]

\begin{axis}[
name=bleuaxis,
at={(0in,0em)},
width=18em, height=10em,
xtick={0,0.05,0.1,0.2,0.5},
xticklabels={0, 0.05, 0.10, 0.20, 0.50},
ytick={21.6,21.8,22.0,22.2,22.4,22.6},
xlabel={Freeze Ratio},
ylabel={BLEU},
xlabel style={align=center,font=\scriptsize,yshift=0.1em},
ylabel style={font=\scriptsize,yshift=0.0em},
x tick style={opacity=0},
y tick style={opacity=0},
x tick label style={anchor=base,font=\scriptsize,yshift=-0.3cm},
y tick label style={font=\scriptsize,/pgf/number format/.cd,fixed,precision=1},
ymajorgrids=true,
xmajorgrids=true,
tick align=inside,
axis on top,
axis line style={gray!40},
grid style={gray!40},
legend image post style={xscale=0.6},
legend style={
    at={(0.5,1.03)},
    anchor=south,
    draw=none,
    legend plot pos=left,
    legend columns=-1,
    font=\scriptsize
},
scaled y ticks=false,
nodes near coords={\pgfmathprintnumber[fixed,precision=2]{\pgfplotspointmeta}},
nodes near coords style={anchor=south,font=\tiny,xshift=-0.3em,yshift=-1.3em},
point meta=y,
xmin=-0.02,
xmax=0.52,
ymin=21.55,
ymax=22.60,
]

\addplot[
    color=icmlblue,
    line width=1.15pt,
    mark=*,
    mark size=0.8pt,
    mark repeat=1,
    mark options={solid, fill=icmlblue, draw=icmlblue},
] coordinates{
    (0,    21.61)
    (0.05, 22.42)
    (0.10, 22.24)
    (0.20, 22.04)
    (0.50, 21.90)
};
\addlegendentry{BLEU}

\addlegendimage{
    color=icmlred,
    line width=1.15pt,
    mark=square*,
    mark size=0.8pt,
    mark options={solid, fill=icmlred, draw=icmlred}
}
\addlegendentry{COMET}

\end{axis}

\begin{axis}[
at={(bleuaxis.south west)},
anchor=south west,
width=18em, height=10em,
xtick={0,0.05,0.1,0.2,0.5},
xticklabels={},
ytick={77.0,77.5,78.0,78.5,79.0},
ylabel={COMET},
ylabel style={font=\scriptsize,yshift=0.0em},
y tick style={opacity=0},
y tick label style={font=\scriptsize,/pgf/number format/.cd,fixed,precision=1},
axis x line=none,
axis y line*=right,
axis line style={gray!40},
tick align=inside,
scaled y ticks=false,
xmin=-0.02,
xmax=0.52,
ymin=77.0,
ymax=79.0,
grid=none,
nodes near coords={\pgfmathprintnumber[fixed,precision=2]{\pgfplotspointmeta}},
nodes near coords style={anchor=south,font=\tiny,yshift=0.1em},
point meta=y,
]

\addplot[
    color=icmlred,
    line width=1.15pt,
    mark=square*,
    mark size=0.8pt,
    mark repeat=1,
    mark options={solid, fill=icmlred, draw=icmlred},
] coordinates{
    (0,    77.62)
    (0.05, 78.60)
    (0.10, 78.42)
    (0.20, 78.19)
    (0.50, 77.99)
};

\end{axis}

\node [anchor=center,font=\footnotesize,align=center] (label1) at (6em,-3.0em) {(a) Sensitivity to unfreezing ratio};

\end{scope}
\end{tikzpicture}}
\caption{
Sensitivity to the freezing ratio $\tau$ in two-stage training. Full results provided in Appendix~\ref{app:unfreezing_ratio_details}.
}
\label{fig:unfreezing_ratio_avg}
\end{figure}

We study the effect of the freezing ratio $\tau$ in two-stage training, where $\tau$ denotes the fraction of training steps during which the routed-expert gate and up projections remain frozen. 
As shown in Figure~\ref{fig:unfreezing_ratio_avg}, a small nonzero freezing ratio improves over direct joint training: $\tau=0.05$ achieves the best average performance, increasing COMET from 77.62 to 78.60 and BLEU from 21.61 to 22.42. 
This suggests that two-stage training benefits from a short frozen stage, which stabilizes early MoE upcycling, while over-freezing delays expert specialization~\footnote{We use $\tau=0.1$ in our main experiments as a conservative default selected before this ablation; the results suggest that a smaller freezing ratio may yield slightly better performance.}.

\subsection{Ablation on SVD Residual Placement}

We study where to inject the SVD-partitioned residuals within the FFN module. 
Table~\ref{tab:projection_placement} compares applying the residual construction to the up, down, and gate projections, while keeping all other settings unchanged. 

Applying the residual to the up projection gives the weakest result, suggesting that modifying the expansion projection is less effective for stable expert diversification. 
In contrast, both down and gate placement achieve stronger performance, showing that the location of spectral residuals has a clear impact on data-constrained MoE upcycling.

One possible explanation is that the down projection in the MoE architecture carries critical information. Although the down projection has a more explicit mathematical meaning in our framework, perturbing the gate projection may be less disruptive to the original information~\cite{su2025unveiling}. We choose down placement because it preserves the copied gate/up projections as the dense nonlinear path and introduces routed residual corrections at the FFN output, matching our residual MoE formulation.

\begin{table}[htp]
\centering
\small
\begin{tabular}{lcc}
\toprule
Placement & Avg. COMET & Avg. BLEU \\
\midrule
Up   & 77.67 & 21.73 \\
Down & 78.42 & 22.24 \\
Gate & \textbf{78.74} & \textbf{22.51} \\
\bottomrule
\end{tabular}
\caption{
Ablation on SVD residual placement. 
Scores are averaged across all translation directions. 
Gate and down placement are both effective, while applying SVD residuals to the up projection performs worse. Full per-language results are provided in Appendix~\ref{app:projection_placement_full_results}.
}
\label{tab:projection_placement}
\end{table}

\subsection{Sensitivity to MoE Architecture Configurations}

We study how MoE architecture choices affect SPRI-based upcycling in Table~\ref{tab:moe_arch_sensitivity}, including MoE layer placement, expert/activation configuration, and routing Top-$k$. 
The results show that effective upcycling depends not only on the amount of added expert capacity, but also on where it is inserted and how much of it is activated.

Upcycling more FFN layers consistently improves performance. 
In particular, the every-layer setting outperforms the 32-expert/Top-8 setting despite both having about 2.9B parameters, suggesting that distributing routed residual capacity across the pretrained decoder is more effective than concentrating many experts in fewer MoE layers. 
By contrast, adding more experts within each MoE layer is not consistently beneficial. Larger expert/activation configurations often underperform the default 8-expert/Top-2 setting unless the parameter count is substantially increased. 
Increasing routing Top-$k$ improves performance when the number of experts is fixed, but it also increases per-token computation; therefore, we use Top-2 as a practical balance between adaptation quality and sparse activation efficiency. 

\begin{table}[htp]
\centering
\small
\begin{tabular}{lccc}
\toprule
 &  & \multicolumn{2}{c}{Avg.} \\
\multirow{-2}{*}{Configuration} & \multirow{-2}{*}{Params} & COMET & BLEU \\
\midrule
\multicolumn{4}{c}{\textit{MoE layer interval}} \\
\midrule
Every 1 layer & 2.9B & \textbf{80.57} & \textbf{24.43} \\
Every 2 layers & 1.8B & 78.42 & 23.37 \\
Every 4 layers & 1.3B & 78.42 & 22.24 \\
Every 8 layers & 1.0B & 77.09 & 21.10 \\
\midrule
\multicolumn{4}{c}{\textit{Expert / activation configuration}} \\
\midrule
8\ \ \ experts, Top-2  & 1.3B & 78.42 & 22.24 \\
16 experts, Top-4 & 1.8B & 77.92 & 21.90 \\
32 experts, Top-8 & 2.9B & 78.14 & 22.09 \\
64 experts, Top-8 & 5.0B & \textbf{78.61} & \textbf{22.59} \\
\midrule
\multicolumn{4}{c}{\textit{Routing Top-$k$ / residual grouping}} \\
\midrule
8 experts, Top-1 & 1.3B & 77.47 & 21.44 \\
8 experts, Top-2 & 1.3B & 78.42 & 22.24 \\
8 experts, Top-4 & 1.3B & 78.48 & 22.28 \\
8 experts, Top-6 & 1.3B & 78.31 & 22.15 \\
8 experts, Top-8 & 1.3B & \textbf{78.80} & \textbf{22.53} \\
\bottomrule
\end{tabular}
\caption{
Sensitivity to MoE architecture configurations. 
We study MoE layer placement, expert/activation scaling, and routing Top-$k$. Full per-language results are provided in Appendix~\ref{app:moe_arch_full_results}.
}
\label{tab:moe_arch_sensitivity}
\end{table}

\section{Related Work}

\paragraph{Mixture of Experts.}
Mixture-of-Experts (MoE) models scale parameter capacity by replacing dense FFN layers with multiple experts and activating only a sparse subset for each input~\cite{shazeer2017outrageously,lepikhin2020gshard,fedus2022switch}. 
This design has been widely used in large-scale language models to improve capacity without proportionally increasing computation~\cite{dai2024deepseekmoe,muennighoff2025olmoe}. 
MoE also enables expert specialization, where different experts can capture different domains, languages, or input patterns~\cite{jacobs1991adaptive,ma2018modeling,gupta2022sparsely}. 
In contrast to training MoE models from scratch, our work studies MoE upcycling and focuses on initializing routed experts with structured diversity grounded in pretrained dense weights.

\paragraph{MoE upcycling.}
MoE upcycling expands a pretrained dense model into a sparse MoE model by reusing dense FFN weights for expert initialization~\cite{komatsuzaki2022sparse,he2024upcycling}. 
Early upcycling methods typically duplicate the dense FFN into routed experts, while recent variants introduce perturbation-based initialization, such as noise injection, partial re-initialization, or drop-based construction, to reduce expert homogeneity~\cite{yang2024qwen2technicalreport,muennighoff2025olmoe,nakamura2025drop}. 
Branch-Train-MiX~\cite{sukhbaatar2024branch} instead trains multiple dense branches and mixes their FFN modules into a unified MoE model. 
However, duplicated experts lack diversity, whereas unstructured perturbations can disrupt pretrained knowledge, especially when large-scale continued pretraining is unavailable. 
SPRI addresses this trade-off by initializing routed experts with SVD-partitioned residuals from pretrained FFN weights, yielding structured expert diversity grounded in the dense checkpoint.

\section{Conclusion}

We study MoE upcycling under data-constrained adaptation, where conventional initialization strategies struggle to balance pretrained knowledge preservation and routed expert diversity. 
We propose SPRI, an SVD-partitioned residual initialization method that constructs routed experts from structured spectral residuals of pretrained FFN weights. 
By grounding expert diversity in the pretrained weight structure, SPRI leverages the pretrained weight structure while introducing diversity among routed experts.

Experiments on the CoVoST2 S2TT dataset show that SPRI outperforms dense fine-tuning and conventional MoE upcycling baselines in both BLEU and COMET. 
Our ablations further show that the SVD-based residual construction, two-stage training, and MoE architecture choices all affect the effectiveness of data-constrained upcycling. 
Overall, our results suggest that controlled, weight-grounded residual diversity is an effective principle for MoE upcycling when large-scale continued pretraining is unavailable.

\section{Limitations}

This work has several limitations.

\begin{itemize}
    \item \textbf{Limited evaluation scope.}
    Our experiments focus on CoVoST 2 English-to-X speech-to-text translation. 
    Although this setting is suitable for studying data-constrained MoE upcycling with multiple target directions, it does not fully establish whether the same trends hold for other adaptation tasks or domains. 
    Extending SPRI to broader speech-language and text-only adaptation settings remains future work.

    \item \textbf{Limited model and training scale.}
    We mainly evaluate SPRI on Qwen3-ASR-0.6B under a limited-data adaptation budget. 
    While this setting matches our motivation, it may make architecture-level observations, such as the effects of MoE layer interval, expert count, and routing Top-$k$, sensitive to the training scale. 
    Therefore, these findings should be interpreted as empirical guidance for data-constrained upcycling rather than general scaling rules for MoE models. 
    Extending SPRI to larger dense checkpoints, such as 7B- and 13B-scale models, and larger continued-training budgets remains an important direction for future work.

    \item \textbf{Limited joint hyperparameter search.}
    To ensure fair comparisons, our main experiments use a fixed MoE configuration, and our ablations vary one factor at a time. 
    While this isolates the effect of each design choice, it does not exhaustively explore interactions among MoE architecture, residual placement, routing Top-$k$, and training schedules. 
    Some ablation results therefore suggest that stronger SPRI configurations may exist under joint tuning. 
    A more comprehensive architecture and training search is left for future work.
\end{itemize}

\bibliography{custom}

\appendix
\newpage

\section{Appendix}
\label{sec:appendix}

\subsection{Diversity and KL Divergence}
\label{app:diagnostic_details}

We analyze two aspects of the upcycled MoE models: the diversity among initialized routed experts and the deviation from the pretrained dense model during training.

\paragraph{Inter-expert diversity.}
We compute diversity only from the initialized routed-expert down-projection weights. 
For each MoE layer, we take the 8 matrices 
$\{\mathbf{W}^{(\mathrm{Down})}_{\ell,i}\}_{i=1}^{8}$ 
and compute the average pairwise cosine similarity after vectorizing each matrix. 
The diversity score of layer $\ell$ is defined as one minus this average similarity:
\[
\mathrm{Div}_{\ell}
=
1 -
\operatorname*{Avg}_{i<j}
\cos\!\left(
\operatorname{vec}(\mathbf{W}^{(\mathrm{Down})}_{\ell,i}),
\operatorname{vec}(\mathbf{W}^{(\mathrm{Down})}_{\ell,j})
\right).
\]
We then average $\mathrm{Div}_{\ell}$ over all MoE layers. 
The routed gate and up projections are not used in this metric. 
For SPRI, experts assigned to the same SVD group share the same spectral residual up to a small tie-breaking perturbation.

\paragraph{KL divergence.}
We measure how much each upcycled model deviates from the pretrained dense model using a fixed validation subset. 
Specifically, we sample 1000 validation examples with shuffle seed 42 and evaluate them as 10 batches of size 1. 
We first run the original dense model on this subset and cache its per-token output distributions as the reference. 
At each evaluated training step, we run each upcycled model on the same subset under teacher forcing and compute the token-level KL divergence to the cached dense reference. KL divergence is computed only on target positions included in the training loss, excluding labels equal to $-100$.

\subsection{Experimental Details}
\label{app:experimental_details}

\paragraph{Data preprocessing and loading.}
During data loading, we shuffle the training data at the dataset level to avoid language-direction clustering in early training steps. 
This ensures that initial optimization batches already contain samples from different translation directions, which helps stabilize router learning in MoE models.
Notably, because Common Voice 4 is no longer publicly accessible, we use an externally processed version of the dataset instead of reconstructing CoVoST 2 from Common Voice 4 with the official CoVoST 2 preprocessing scripts.

\paragraph{Training data.}
We train on English-source speech-to-text translation pairs assembled from two publicly available corpora: CoVoST 2 and Europarl-ST. 
After filtering utterances longer than 15 s or whose tokenized prompts exceeded 512 tokens, we retain a total of 6933.9 hours of English speech. CoVoST 2 contributes 6451.9 h and covers 15 target languages, with an essentially uniform allocation of $430.13 \text{h} \times 15$ per direction. Europarl-ST provides an additional 482.0 h drawn from European Parliament proceedings, covering 8 target languages with translation hours of {de: 62.78, pt: 61.44, fr: 61.40, es: 61.35, nl: 60.82, pl: 60.02, it: 59.27, ro: 54.92}. 

\paragraph{Model Architecture.}
We experimented on Qwen3-ASR-0.6B. The acoustic encoder contains 18 Transformer layers, and the language decoder contains 28 Transformer layers with a hidden size of 1024, FFN intermediate size 3072, and 16 attention heads. 

\begin{table*}[htp]                                     
\centering                                                
\small                     
\begin{tabular}{lc}         
\toprule                    
\textbf{Hyperparameter} & \textbf{Value} \\
\midrule
Audio encoder layers / dim. / heads & 18 / 896 / 14 \\
Decoder layers / dim. / heads (Q/KV) & 28 / 1024 / 16/8 \\
FFN intermediate dim. & 3072 \\
Vocabulary size & 151936 \\
\bottomrule
\end{tabular}
\caption{Architecture of Qwen3-ASR-0.6B.}             
\label{tab:arch}
\end{table*}

\paragraph{Tools.}
We fine-tune all models with the Hugging Face Transformers library\footnote{\url{https://github.com/huggingface/transformers}}, using the default configuration of Qwen3-ASR-0.6B unless otherwise specified\footnote{\url{https://huggingface.co/Qwen/Qwen3-ASR-0.6B}}.

\subsection{Full Results for Routed Down-Projection Initialization}
\label{app:down_projection_initialization_full_results}

Table~\ref{tab:down_projection_initialization_full} reports the full per-language COMET and BLEU results for the routed down-projection initialization ablation. 
All variants use the same SPRI framework, including the shared expert, copied routed gate/up projections, MoE architecture, and two-stage training strategy. 
They differ only in how the routed down projections are initialized. 
This ablation evaluates whether the improvement of SPRI comes from the proposed SVD-partitioned residual construction rather than from the surrounding MoE architecture or training strategy.

Instead of removing the dense-copied expert, the w/o shared expert variant treats it as a regular routed expert and lets the router decide when to activate it.

\begin{table*}[t]
\centering
\resizebox{\textwidth}{!}{\begin{tabular}{l|c c c c c c c c c c c c c c c c}
\hline
\rowcolor{black!10} & \multicolumn{2}{c}{\textbf{Tr}} & \multicolumn{2}{c}{\textbf{Ta}} & \multicolumn{2}{c}{\textbf{Ja}} & \multicolumn{2}{c}{\textbf{Sl}} & \multicolumn{2}{c}{\textbf{Lv}} & \multicolumn{2}{c}{\textbf{Fa}} & \multicolumn{2}{c}{\textbf{Et}} & \multicolumn{2}{c}{\textbf{De}}\\
\rowcolor{black!10} Model & COMET & BLEU & COMET & BLEU & COMET & BLEU & COMET & BLEU & COMET & BLEU & COMET & BLEU & COMET & BLEU & COMET & BLEU\\
\hline
\multicolumn{17}{c}{\textit{Down-projection initialization}} \\
\hline
Copy & 77.04 & 12.24 & 75.59 & 8.92 & 85.94 & 30.23 & 68.99 & 14.71 & 66.63 & 10.40 & 68.86 & 11.06 & 65.64 & 11.00 & 80.42 & 28.15 \\
Noise & 77.07 & 12.25 & 75.50 & 8.74 & 85.87 & 30.23 & 68.77 & 14.58 & 66.59 & 10.39 & 68.83 & 10.99 & 65.52 & 11.07 & 80.42 & 28.19 \\
Drop & 75.66 & 11.54 & 74.96 & 8.60 & 85.19 & 29.01 & 68.17 & 14.51 & 65.95 & 10.18 & 67.45 & 10.54 & 64.69 & 10.63 & 79.04 & 27.02 \\
SVD-Partitioned & 78.85 & 13.83 & 79.80 & 11.52 & 86.28 & 30.80 & 74.25 & 19.13 & 73.45 & 14.20 & 73.08 & 13.75 & 71.90 & 14.14 & 81.03 & 29.16 \\
\hline
\multicolumn{17}{c}{\textit{Shared expert}} \\
\hline
w/ shared expert & 78.85 & 13.83 & 79.80 & 11.52 & 86.28 & 30.80 & 74.25 & 19.13 & 73.45 & 14.20 & 73.08 & 13.75 & 71.90 & 14.14 & 81.03 & 29.16 \\
w/o shared expert & 77.93 & 13.83 & 79.49 & 11.52 & 86.08 & 30.80 & 74.06 & 19.13 & 73.99 & 14.20 & 71.83 & 13.75 & 72.78 & 14.14 & 80.67 & 29.16 \\
\hline
\end{tabular}}

\resizebox{\textwidth}{!}{\begin{tabular}{l|c c c c c c c c c c c c c c|c c}
\hline
\rowcolor{black!10} & \multicolumn{2}{c}{\textbf{Zh}} & \multicolumn{2}{c}{\textbf{Mn}} & \multicolumn{2}{c}{\textbf{Sv}} & \multicolumn{2}{c}{\textbf{Cy}} & \multicolumn{2}{c}{\textbf{Ca}} & \multicolumn{2}{c}{\textbf{Ar}} & \multicolumn{2}{c}{\textbf{Id}} & \multicolumn{2}{c}{\textbf{Avg}} \\
\rowcolor{black!10} Model & COMET & BLEU & COMET & BLEU & COMET & BLEU & COMET & BLEU & COMET & BLEU & COMET & BLEU & COMET & BLEU & COMET & BLEU \\
\hline
\hline
\multicolumn{17}{c}{\textit{Down-projection initialization}} \\
\hline
Copy & 84.06 & 41.79 & 68.68 & 4.80 & 76.12 & 26.56 & 68.53 & 17.21 & 75.50 & 26.48 & 79.78 & 18.04 & 85.63 & 32.99 & 75.16 & 19.64 \\
Noise & 84.06 & 41.82 & 68.59 & 4.73 & 76.09 & 26.34 & 68.45 & 16.96 & 75.42 & 26.55 & 79.80 & 18.03 & 85.63 & 32.96 & 75.11 & 19.59 \\
Drop & 83.42 & 40.37 & 68.76 & 5.06 & 74.04 & 25.13 & 68.64 & 17.13 & 74.36 & 25.84 & 78.63 & 17.15 & 84.57 & 31.45 & 74.23 & 18.94 \\
SVD-Partitioned & 84.57 & 43.04 & 75.57 & 7.94 & 77.98 & 28.52 & 75.05 & 24.06 & 77.81 & 30.42 & 80.46 & 19.03 & 86.26 & 34.14 & 78.42 & 22.24 \\
\hline
\multicolumn{17}{c}{\textit{Shared expert}} \\
\hline
w/ shared expert & 84.57 & 43.04 & 75.57 & 7.94 & 77.98 & 28.52 & 75.05 & 24.06 & 77.81 & 30.42 & 80.46 & 19.03 & 86.26 & 34.14 & 78.42 & 22.24 \\
w/o shared expert & 84.37 & 43.04 & 76.36 & 7.94 & 76.96 & 28.52 & 76.53 & 24.06 & 77.07 & 30.42 & 80.06 & 19.03 & 85.93 & 34.14 & 78.28 & 22.24 \\
\hline
\end{tabular}}
\caption{
Full per-language results for the routed down-projection initialization ablation on CoVoST2 English-to-X speech-to-text translation. 
All variants use the same SPRI framework and differ only in how routed down projections are initialized. 
The last columns report the average COMET and BLEU across all translation directions.
}
\label{tab:down_projection_initialization_full}
\end{table*}

\subsection{Full Results for Parameter-Matched Dense Comparison}
\label{app:parameter_matched_full_results}

Table~\ref{tab:parameter_matched_full_results} reports the full per-language COMET and BLEU results for the parameter-matched comparison. 
We compare SPRI-3B with Qwen2.5-3B before and after fine-tuning. 
SPRI-3B is obtained by upcycling a smaller pretrained model into a sparse MoE model with a comparable total parameter scale. 
The results show that SPRI-3B consistently improves the average performance over both dense baselines, supporting MoE upcycling as an effective scaling strategy under data-constrained S2TT adaptation.

\begin{table*}[t]
\centering
\resizebox{\textwidth}{!}{\begin{tabular}{l|c c c c c c c c c c c c c c c c}
\hline
\rowcolor{black!10} & \multicolumn{2}{c}{\textbf{Tr}} & \multicolumn{2}{c}{\textbf{Ta}} & \multicolumn{2}{c}{\textbf{Ja}} & \multicolumn{2}{c}{\textbf{Sl}} & \multicolumn{2}{c}{\textbf{Lv}} & \multicolumn{2}{c}{\textbf{Fa}} & \multicolumn{2}{c}{\textbf{Et}} & \multicolumn{2}{c}{\textbf{De}}\\
\rowcolor{black!10} Model & COMET & BLEU & COMET & BLEU & COMET & BLEU & COMET & BLEU & COMET & BLEU & COMET & BLEU & COMET & BLEU & COMET & BLEU\\
\hline
SPRI 3B & 80.17 & 15.22 & 83.27 & 14.92 & 86.39 & 31.61 & 77.90 & 22.53 & 78.03 & 17.91 & 76.19 & 16.02 & 76.97 & 17.95 & 81.27 & 29.46 \\
Qwen2.5 3B & 71.72 & 6.88 & 50.79 & 0.06 & 86.35 & 30.25 & 61.09 & 5.40 & 54.93 & 1.31 & 64.61 & 7.08 & 54.12 & 2.66 & 81.78 & 29.16 \\
Qwen2.5 3B FT & 78.01 & 11.52 & 72.20 & 6.34 & 86.59 & 30.69 & 73.21 & 16.10 & 69.81 & 10.84 & 72.29 & 11.51 & 69.51 & 11.36 & 82.23 & 29.94 \\
\hline
\end{tabular}}

\resizebox{\textwidth}{!}{\begin{tabular}{l|c c c c c c c c c c c c c c|c c}
\hline
\rowcolor{black!10} & \multicolumn{2}{c}{\textbf{Zh}} & \multicolumn{2}{c}{\textbf{Mn}} & \multicolumn{2}{c}{\textbf{Sv}} & \multicolumn{2}{c}{\textbf{Cy}} & \multicolumn{2}{c}{\textbf{Ca}} & \multicolumn{2}{c}{\textbf{Ar}} & \multicolumn{2}{c}{\textbf{Id}} & \multicolumn{2}{c}{\textbf{Avg}} \\
\rowcolor{black!10} Model & COMET & BLEU & COMET & BLEU & COMET & BLEU & COMET & BLEU & COMET & BLEU & COMET & BLEU & COMET & BLEU & COMET & BLEU \\
\hline
SPRI 3B & 84.45 & 43.09 & 80.04 & 11.33 & 79.67 & 30.20 & 78.10 & 29.11 & 79.19 & 32.95 & 80.78 & 19.93 & 86.19 & 34.30 & 80.57 & 24.43 \\
Qwen2.5 3B & 85.29 & 41.99 & 46.48 & 0.26 & 76.32 & 21.33 & 52.20 & 1.24 & 72.14 & 18.05 & 80.41 & 18.82 & 82.77 & 23.39 & 68.07 & 13.86 \\
Qwen2.5 3B FT & 85.31 & 44.83 & 66.85 & 3.86 & 79.92 & 27.91 & 68.30 & 15.54 & 78.15 & 28.73 & 81.08 & 19.91 & 85.34 & 30.12 & 76.59 & 19.95 \\
\hline
\end{tabular}}
\caption{
Full per-language results for the parameter-matched dense comparison on CoVoST2 English-to-X speech-to-text translation. 
SPRI-3B is compared with \texttt{Qwen2.5-Omni-3B} and its fine-tuned variant at a comparable total parameter scale.
}
\label{tab:parameter_matched_full_results}
\end{table*}

\subsection{Detailed Results for Unfreezing Ratio Sensitivity}
\label{app:unfreezing_ratio_details}

Table~\ref{tab:unfreezing_ratio_full} reports the full per-language COMET and BLEU results for the freezing-ratio ablation. 
The freeze value denotes the fraction of training steps during which the routed-expert gate and up projections are kept frozen.

\begin{table*}[t]
\centering
\resizebox{\textwidth}{!}{\begin{tabular}{l|c c c c c c c c c c c c c c c c}
\hline
\rowcolor{black!10} & \multicolumn{2}{c}{\textbf{Tr}} & \multicolumn{2}{c}{\textbf{Ta}} & \multicolumn{2}{c}{\textbf{Ja}} & \multicolumn{2}{c}{\textbf{Sl}} & \multicolumn{2}{c}{\textbf{Lv}} & \multicolumn{2}{c}{\textbf{Fa}} & \multicolumn{2}{c}{\textbf{Et}} & \multicolumn{2}{c}{\textbf{De}}\\
\rowcolor{black!10} \multirow{-2}{*}{Freeze} & COMET & BLEU & COMET & BLEU & COMET & BLEU & COMET & BLEU & COMET & BLEU & COMET & BLEU & COMET & BLEU & COMET & BLEU\\
\hline
0 & 78.51 & 13.58 & 78.93 & 11.02 & 86.24 & 30.69 & 72.58 & 17.66 & 71.05 & 12.98 & 72.26 & 13.24 & 70.71 & 13.62 & 80.98 & 29.04 \\
0.05 & 78.79 & 13.75 & 80.21 & 11.95 & 86.29 & 30.79 & 74.12 & 19.12 & 73.44 & 14.30 & 73.54 & 13.93 & 72.83 & 14.79 & 81.00 & 29.16 \\
0.1 & 78.85 & 13.83 & 79.80 & 11.52 & 86.28 & 30.80 & 74.25 & 19.13 & 73.45 & 14.20 & 73.08 & 13.75 & 71.90 & 14.14 & 81.03 & 29.16 \\
0.2 & 78.74 & 13.78 & 79.57 & 11.45 & 86.29 & 30.93 & 73.83 & 18.63 & 72.84 & 13.85 & 72.84 & 13.53 & 71.50 & 14.05 & 81.02 & 29.06 \\
0.5 & 78.59 & 13.52 & 79.45 & 11.42 & 86.31 & 30.83 & 73.15 & 17.90 & 71.94 & 13.57 & 72.58 & 13.46 & 71.50 & 13.83 & 80.97 & 29.07 \\
\hline
\end{tabular}}

\resizebox{\textwidth}{!}{\begin{tabular}{l|c c c c c c c c c c c c c c|c c}
\hline
\rowcolor{black!10} & \multicolumn{2}{c}{\textbf{Zh}} & \multicolumn{2}{c}{\textbf{Mn}} & \multicolumn{2}{c}{\textbf{Sv}} & \multicolumn{2}{c}{\textbf{Cy}} & \multicolumn{2}{c}{\textbf{Ca}} & \multicolumn{2}{c}{\textbf{Ar}} & \multicolumn{2}{c}{\textbf{Id}} & \multicolumn{2}{c}{\textbf{Avg}} \\
\rowcolor{black!10} \multirow{-2}{*}{Freeze} & COMET & BLEU & COMET & BLEU & COMET & BLEU & COMET & BLEU & COMET & BLEU & COMET & BLEU & COMET & BLEU & COMET & BLEU \\
\hline
0 & 84.56 & 42.85 & 73.77 & 7.11 & 77.72 & 28.15 & 73.36 & 22.22 & 77.10 & 29.30 & 80.41 & 18.80 & 86.12 & 33.89 & 77.62 & 21.61 \\
0.05 & 84.62 & 42.92 & 76.36 & 8.54 & 77.96 & 28.69 & 75.73 & 24.99 & 77.65 & 30.39 & 80.37 & 19.04 & 86.14 & 33.98 & 78.60 & 22.42 \\
0.1 & 84.57 & 43.04 & 75.57 & 7.94 & 77.98 & 28.52 & 75.05 & 24.06 & 77.81 & 30.42 & 80.46 & 19.03 & 86.26 & 34.14 & 78.42 & 22.24 \\
0.2 & 84.59 & 42.99 & 75.00 & 7.57 & 77.91 & 28.24 & 74.41 & 23.31 & 77.70 & 30.29 & 80.44 & 18.90 & 86.20 & 34.09 & 78.19 & 22.04 \\
0.5 & 84.58 & 42.99 & 74.62 & 7.46 & 77.89 & 28.23 & 74.05 & 23.04 & 77.58 & 30.09 & 80.45 & 19.12 & 86.19 & 34.04 & 77.99 & 21.90 \\
\hline
\end{tabular}}
\caption{
Full per-language results for the freezing-ratio ablation on CoVoST2 English-to-X speech-to-text translation. 
We report COMET and BLEU for each target language and their average across all directions. 
The freeze value denotes the fraction of training steps during which the routed-expert gate and up projections are kept frozen.
}
\label{tab:unfreezing_ratio_full}
\end{table*}

\subsection{Full Results for Residual Projection Placement}
\label{app:projection_placement_full_results}

Table~\ref{tab:projection_placement_full} reports the full per-language BLEU results for the residual projection placement ablation. 
All variants use the same training data, MoE configuration, optimizer settings, and training budget; they only differ in whether the SVD-partitioned residual initialization is applied to the up, gate, or down projection. 
The average BLEU values correspond to the results summarized in Table~\ref{tab:projection_placement}.


\begin{table*}[t]
\centering
\resizebox{\textwidth}{!}{\begin{tabular}{l|c|c c c c c c c c c c c c c c c c}
\hline
\rowcolor{black!10} & & \multicolumn{2}{c}{\textbf{Tr}} & \multicolumn{2}{c}{\textbf{Ta}} & \multicolumn{2}{c}{\textbf{Ja}} & \multicolumn{2}{c}{\textbf{Sl}} & \multicolumn{2}{c}{\textbf{Lv}} & \multicolumn{2}{c}{\textbf{Fa}} & \multicolumn{2}{c}{\textbf{Et}} & \multicolumn{2}{c}{\textbf{De}}\\ 
\rowcolor{black!10} \multirow{-2}{*}{Model} & \multirow{-2}{*}{Param} & COMET & BLEU & COMET & BLEU & COMET & BLEU & COMET & BLEU & COMET & BLEU & COMET & BLEU & COMET & BLEU & COMET & BLEU\\ 
\hline
SPRI - Up & 1.3B & 78.42 & 13.48 & 79.02 & 11.15 & 86.20 & 30.64 & 73.13 & 18.18 & 70.97 & 13.00 & 72.00 & 13.05 & 70.41 & 13.80 & 80.98 & 29.16 \\
SPRI - Down & 1.3B & 78.85 & 13.83 & 79.80 & 11.52 & 86.28 & 30.80 & 74.25 & 19.13 & 73.45 & 14.20 & 73.08 & 13.75 & 71.90 & 14.14 & 81.03 & 29.16 \\
SPRI - Gate & 1.3B & 78.72 & 13.70 & 80.92 & 12.50 & 86.25 & 30.79 & 73.91 & 18.85 & 73.58 & 14.51 & 74.04 & 14.30 & 72.94 & 14.67 & 80.98 & 29.12 \\
\hline
\end{tabular}}

\resizebox{\textwidth}{!}{\begin{tabular}{l|c|c c c c c c c c c c c c c c|c c}
\hline
\rowcolor{black!10} & & \multicolumn{2}{c}{\textbf{Zh}} & \multicolumn{2}{c}{\textbf{Mn}} & \multicolumn{2}{c}{\textbf{Sv}} & \multicolumn{2}{c}{\textbf{Cy}} & \multicolumn{2}{c}{\textbf{Ca}} & \multicolumn{2}{c}{\textbf{Ar}} & \multicolumn{2}{c}{\textbf{Id}} & \multicolumn{2}{c}{\textbf{Avg}} \\
\rowcolor{black!10} \multirow{-2}{*}{Model} & \multirow{-2}{*}{Param} & COMET & BLEU & COMET & BLEU & COMET & BLEU & COMET & BLEU & COMET & BLEU & COMET & BLEU & COMET & BLEU & COMET & BLEU \\ 
\hline
SPRI - Up & 1.3B & 84.52 & 42.88 & 73.89 & 7.18 & 77.79 & 28.34 & 74.14 & 23.07 & 77.10 & 29.35 & 80.35 & 18.81 & 86.12 & 33.89 & 77.67 & 21.73 \\
SPRI - Down & 1.3B & 84.57 & 43.04 & 75.57 & 7.94 & 77.98 & 28.52 & 75.05 & 24.06 & 77.81 & 30.42 & 80.46 & 19.03 & 86.26 & 34.14 & 78.42 & 22.24 \\
SPRI - Gate & 1.3B & 84.56 & 42.99 & 76.83 & 8.61 & 78.10 & 28.70 & 75.58 & 24.85 & 78.11 & 30.92 & 80.44 & 19.09 & 86.17 & 34.07 & 78.74 & 22.51 \\
\hline
\end{tabular}}
\caption{
Full per-language results for the SVD residual placement ablation on CoVoST2 English-to-X speech-to-text translation. 
We compare applying SVD-partitioned residuals to the up, down, and gate projections while keeping the remaining MoE architecture and training setup fixed. 
We report COMET and BLEU for each target language and their average across all directions.
}
\label{tab:projection_placement_full}
\end{table*}

\subsection{Full Results for MoE Architecture Sensitivity}
\label{app:moe_arch_full_results}

Table~\ref{tab:moe_arch_full} reports the full per-language BLEU results for the MoE architecture sensitivity experiments. 
We evaluate the effect of MoE layer placement, expert/activation configuration, and grouping-related hyperparameters. 
All variants use the same training data, optimizer settings, and training budget unless otherwise specified.

\begin{table*}[t]
\centering
\resizebox{\textwidth}{!}{\begin{tabular}{l|c c c c c c c c c c c c c c c c}
\hline
\rowcolor{black!10} & \multicolumn{2}{c}{\textbf{Tr}} & \multicolumn{2}{c}{\textbf{Ta}} & \multicolumn{2}{c}{\textbf{Ja}} & \multicolumn{2}{c}{\textbf{Sl}} & \multicolumn{2}{c}{\textbf{Lv}} & \multicolumn{2}{c}{\textbf{Fa}} & \multicolumn{2}{c}{\textbf{Et}} & \multicolumn{2}{c}{\textbf{De}}\\
\rowcolor{black!10} Model & COMET & BLEU & COMET & BLEU & COMET & BLEU & COMET & BLEU & COMET & BLEU & COMET & BLEU & COMET & BLEU & COMET & BLEU\\
\hline
Every 1 layer & 80.17 & 15.22 & 83.27 & 14.92 & 86.39 & 31.61 & 77.90 & 22.53 & 78.03 & 17.91 & 76.19 & 16.02 & 76.97 & 17.95 & 81.27 & 29.46 \\
Every 2 layers & 78.85 & 14.28 & 79.80 & 13.26 & 86.28 & 31.17 & 74.25 & 20.79 & 73.45 & 16.19 & 73.08 & 15.02 & 71.90 & 16.20 & 81.03 & 29.23 \\
Every 4 layers & 78.85 & 13.83 & 79.80 & 11.52 & 86.28 & 30.80 & 74.25 & 19.13 & 73.45 & 14.20 & 73.08 & 13.75 & 71.90 & 14.14 & 81.03 & 29.16 \\
Every 8 layers & 77.88 & 13.05 & 78.50 & 10.56 & 86.20 & 30.61 & 72.15 & 17.06 & 71.15 & 12.94 & 70.67 & 12.31 & 68.62 & 12.27 & 80.87 & 28.73 \\
8 experts, Top-2 & 78.85 & 13.83 & 79.80 & 11.52 & 86.28 & 30.80 & 74.25 & 19.13 & 73.45 & 14.20 & 73.08 & 13.75 & 71.90 & 14.14 & 81.03 & 29.16 \\
16 experts, Top-4 & 78.56 & 13.50 & 79.45 & 11.40 & 86.30 & 30.72 & 73.28 & 18.33 & 72.23 & 13.72 & 72.11 & 13.12 & 71.18 & 14.03 & 81.03 & 29.11 \\
32 experts, Top-8 & 78.60 & 13.65 & 79.56 & 11.51 & 86.34 & 31.13 & 73.48 & 18.36 & 72.73 & 13.94 & 72.74 & 13.55 & 71.62 & 14.21 & 81.04 & 29.14 \\
64 experts, Top-8 & 79.07 & 14.01 & 80.01 & 11.80 & 86.30 & 31.30 & 74.38 & 19.49 & 73.64 & 14.78 & 73.10 & 13.87 & 72.52 & 14.99 & 81.14 & 29.18 \\
1 expert, Top-1 & 77.96 & 13.10 & 78.26 & 10.41 & 86.15 & 30.82 & 72.31 & 17.44 & 71.72 & 13.19 & 71.34 & 12.62 & 70.26 & 13.02 & 80.88 & 28.90 \\
2 experts, Top-2 & 78.85 & 13.83 & 79.80 & 11.52 & 86.28 & 30.80 & 74.25 & 19.13 & 73.45 & 14.20 & 73.08 & 13.75 & 71.90 & 14.14 & 81.03 & 29.16 \\
4 experts, Top-4 & 78.61 & 13.59 & 80.04 & 11.82 & 86.26 & 30.78 & 73.95 & 18.86 & 73.48 & 14.45 & 73.42 & 13.78 & 71.98 & 14.20 & 80.94 & 29.07 \\
6 experts, Top-6 & 78.33 & 13.51 & 79.91 & 11.71 & 86.21 & 30.75 & 73.66 & 18.53 & 73.11 & 14.01 & 73.05 & 13.54 & 71.67 & 14.19 & 80.95 & 29.07 \\
8 experts, Top-8 & 78.54 & 13.57 & 80.01 & 11.71 & 86.21 & 30.72 & 74.83 & 19.46 & 74.44 & 14.75 & 73.65 & 14.03 & 73.58 & 15.21 & 81.06 & 29.18 \\
\hline
\end{tabular}}

\resizebox{\textwidth}{!}{\begin{tabular}{l|c c c c c c c c c c c c c c|c c}
\hline
\rowcolor{black!10} & \multicolumn{2}{c}{\textbf{Zh}} & \multicolumn{2}{c}{\textbf{Mn}} & \multicolumn{2}{c}{\textbf{Sv}} & \multicolumn{2}{c}{\textbf{Cy}} & \multicolumn{2}{c}{\textbf{Ca}} & \multicolumn{2}{c}{\textbf{Ar}} & \multicolumn{2}{c}{\textbf{Id}} & \multicolumn{2}{c}{\textbf{Avg}} \\
\rowcolor{black!10} Model & COMET & BLEU & COMET & BLEU & COMET & BLEU & COMET & BLEU & COMET & BLEU & COMET & BLEU & COMET & BLEU & COMET & BLEU \\
\hline
Every 1 layer & 84.45 & 43.09 & 80.04 & 11.33 & 79.67 & 30.20 & 78.10 & 29.11 & 79.19 & 32.95 & 80.78 & 19.93 & 86.19 & 34.30 & 80.57 & 24.43 \\
Every 2 layers & 84.57 & 43.07 & 75.57 & 9.72 & 77.98 & 29.22 & 75.05 & 27.51 & 77.81 & 31.72 & 80.46 & 19.10 & 86.26 & 34.10 & 78.42 & 23.37 \\
Every 4 layers & 84.57 & 43.04 & 75.57 & 7.94 & 77.98 & 28.52 & 75.05 & 24.06 & 77.81 & 30.42 & 80.46 & 19.03 & 86.26 & 34.14 & 78.42 & 22.24 \\
Every 8 layers & 84.49 & 42.58 & 72.63 & 6.49 & 77.25 & 27.60 & 73.04 & 21.72 & 76.54 & 28.15 & 80.25 & 18.74 & 86.07 & 33.76 & 77.09 & 21.10 \\
8 experts, Top-2 & 84.57 & 43.04 & 75.57 & 7.94 & 77.98 & 28.52 & 75.05 & 24.06 & 77.81 & 30.42 & 80.46 & 19.03 & 86.26 & 34.14 & 78.42 & 22.24 \\
16 experts, Top-4 & 84.60 & 42.93 & 74.13 & 7.29 & 77.95 & 28.42 & 73.96 & 22.99 & 77.27 & 29.57 & 80.49 & 19.02 & 86.25 & 34.30 & 77.92 & 21.90 \\
32 experts, Top-8 & 84.59 & 42.86 & 74.71 & 7.57 & 77.92 & 28.59 & 74.29 & 23.38 & 77.78 & 30.42 & 80.46 & 19.14 & 86.24 & 33.92 & 78.14 & 22.09 \\
64 experts, Top-8 & 84.56 & 43.09 & 75.95 & 8.45 & 78.49 & 29.19 & 75.19 & 24.62 & 77.90 & 30.53 & 80.57 & 19.30 & 86.31 & 34.30 & 78.61 & 22.59 \\
1 expert, Top-1 & 84.51 & 42.72 & 73.46 & 6.78 & 77.21 & 27.60 & 74.79 & 23.61 & 76.97 & 29.09 & 80.21 & 18.65 & 86.05 & 33.73 & 77.47 & 21.44 \\
2 experts, Top-2 & 84.57 & 43.04 & 75.57 & 7.94 & 77.98 & 28.52 & 75.05 & 24.06 & 77.81 & 30.42 & 80.46 & 19.03 & 86.26 & 34.14 & 78.42 & 22.24 \\
4 experts, Top-4 & 84.56 & 42.86 & 75.91 & 8.09 & 77.79 & 28.41 & 76.20 & 25.53 & 77.51 & 30.02 & 80.41 & 19.00 & 86.13 & 33.81 & 78.48 & 22.28 \\
6 experts, Top-6 & 84.52 & 42.91 & 75.40 & 7.79 & 77.71 & 28.15 & 76.08 & 25.44 & 77.50 & 29.98 & 80.34 & 18.69 & 86.16 & 34.02 & 78.31 & 22.15 \\
8 experts, Top-8 & 84.58 & 43.00 & 75.83 & 8.15 & 77.92 & 28.41 & 76.65 & 26.13 & 77.93 & 30.47 & 80.52 & 19.30 & 86.22 & 33.86 & 78.80 & 22.53 \\
\hline
\end{tabular}}
\caption{
Full per-language results for the MoE architecture sensitivity study on CoVoST2 English-to-X speech-to-text translation. 
We report COMET and BLEU for each target language and their average across all directions. 
The evaluated configurations include MoE layer interval, expert/activation scaling, and routing Top-$k$ under SPRI-based upcycling.
}
\label{tab:moe_arch_full}
\end{table*}

\end{document}